\documentclass[runningheads]{llncs}
\usepackage[T1]{fontenc}
\usepackage{graphicx}
\usepackage{color}
\usepackage[hidelinks]{hyperref}

\usepackage{amsmath} 

\usepackage[linesnumbered, ruled, vlined]{algorithm2e}  

\usepackage{footmisc}
\usepackage{longtable}
\usepackage{booktabs}
\usepackage{amsfonts}
\usepackage{setspace}
\usepackage{subcaption}
\usepackage{graphicx}
\usepackage{caption}

\begin{document}

\title{SLAP: Stratified Loss-based Pruning for On-Policy Data-Efficient Instruction Tuning}
%
%
\author{Run Zou\inst{1} \and
Jianhang Ding\inst{1} \and
Yifan Ding\inst{2} \and Wen Wu\inst{2} \and Hao Chen\inst{2} \\
\and Renshu Gu\inst{3}\textsuperscript{*}}
\authorrunning{F. Author et al.}
%
\institute{Alibaba International Digital Commerce Group, HangZhou, China \and
Hangzhou Lingju Intelligence AI Lab, Hangzhou, China \and
Hangzhou Dianzi University, HangZhou, China\\
\email{renshugu@hdu.edu.cn}}

\maketitle              
\begin{abstract}
Instruction tuning has optimized the specialized capabilities of large language models (LLMs), but it often requires extensive datasets and prolonged training times. The challenge lies in developing specific capabilities by identifying useful data and efficiently fine-tuning. High-quality and diverse pruned data can help models achieve lossless performance at a lower cost. In this paper, we propose \textbf{SLAP}, a novel batch-aware data selection framework that evaluates the learnability of entire batch compositions rather than individual. SLAP ensures comprehensive data distribution coverage through distribution-aware stratified sampling while maximizing intra-batch diversity through relative distance optimization. By leveraging Hessian-approximated gradient information for dynamic batch selection, SLAP significantly outperforms existing state-of-the-art methods across multiple model architectures (LLaMA, ChatGLM) and diverse downstream tasks including multi-turn dialogue, multilingual translation, and question answering. Most notably, SLAP achieves superior performance with 20-40\% less training data compared to full dataset training, substantially reducing computational costs while maintaining or improving model capabilities. These results establish SLAP as a powerful approach for efficient and effective instruction tuning of large language models.

\end{abstract}
\section{Introduction}
Instructional tuning has become essential for enhancing LLM capabilities\cite{ouyang2022training}. While recent research focuses on high-quality data collection and on-policy training strategies\cite{hong2024diversified,xia2024less,everaert2023gio}, the challenge of data quality persists, as duplicate and low-quality data can degrade model performance\cite{hernandez2022scaling}.

Current data selection methods fall into two categories: off-policy and on-policy approaches\cite{zheng2022coverage,hong2024diversified,zheng2024elfs}. Off-policy methods rely on static features like loss\cite{jiang2018mentornet,wei2020combating}, influence scores\cite{koh2017understanding,yang2022dataset}, or embedding-based metrics. However, these methods lack adaptability to model updates. On-policy methods\cite{mindermann2022prioritized,deng2023towards,hong2024diversified,paul2021deep,pooladzandi2022adaptive} calculate importance scores in real-time but require substantial computational resources\cite{mindermann2022prioritized}. While Feng\cite{hong2024diversified} improved batch selection through orthogonal representativeness, data learnability remains unexplored.

This paper proposes an on-policy data selection framework, dubbed SLAP, that considers batch learnability, data coverage, data diversity, and computational efficiency.
SLAP evaluates the learnability of entire batch compositions rather than individual. To achieve comprehensive data distribution coverage, SLAP approximates the NP-hard global search in coreset selection by distribution-aware stratified sampling. We provide a theoretical analysis for coreset selection in Natural Language Processing (NLP) tasks from the perspective of geometric coverage. Meanwhile, SLAP maximizes intra-batch diversity through relative distance optimization. This prevents selecting samples that contain redundant information\cite{hong2024diversified} and increases the learnability of the data. Inspired by the Adam algorithm\cite{kingma2017adammethodstochasticoptimization}, we integrate second-moment cumulative gradient updates to reduce fluctuations from random sampling, helping the model to recognize key features across batches consistently.

Through extensive experiments, we demonstrate that SLAP achieves optimal loss across various pruning rates and LLMs (LLaMa3, ChatGLM3), consistently maintaining or improving performance while reducing computational costs by 20-40\%. Our results show particular strength in handling multi-turn dialogues, multilingual translation, and complex question-answering tasks, suggesting broad applicability across different Natural Language Processing (NLP) domains. Furthermore, SLAP exhibits robust generalization capabilities, maintaining consistent performance even with reduced training data, making it particularly valuable for resource-constrained scenarios.

Our contributions can be summarized as follows:
\begin{enumerate}
    \item  We propose an on-policy batch-aware data pruning strategy that preserves data coverage through distribution-aware stratified sampling 
    and maximizes data diversity within batches. 
    \item  We propose a Hessian-approximated gradient optimization method to maximize sample distance in high-dimensional feature space, which enables more precision and dynamics than embeddings.
    \item We provide a theoretical analysis for coreset selection in instructional-tuning of large language models. Further, we show how to solve the NP-hard problem in practice using an efficient approximation.
    \item We evaluate our approach on three diverse downstream datasets: llama3-Chinese-chat (LLaMaQA), WikiMatrix, and our net literature dialogue dataset (NetLit). Results demonstrate that SLAP consistently achieves superior performance in various pruning rates and LLMs (LLaMa3 and ChatGLM3) with lower computational cost. 
\end{enumerate}

\section{Related work}
\textbf{Coreset selection.} Existing methods\cite{guo2022deepcore,yoon2021online} focus on creating representative data subsets for efficient training. While traditional approaches prioritize difficult samples\cite{sorscher2022beyond}, this can lead to including outliers and noise. Although\cite{xia2022moderate} addresses this by selecting median-difficulty samples, their method lacks consideration of data diversity. Zheng's\cite{zheng2022coverage} stratified approach with K strata improves distribution coverage but fails to guarantee the learning value of selected samples.

\textbf{On-policy batch selection.}
Recent approaches fall into two categories: dependent on the reference models\cite{evans2024datacurationjointexample,mindermann2022prioritizedtrainingpointslearnable} and not dependent\cite{hong2024diversified,qin2023infobatchlosslesstrainingspeed}. While Feng\cite{hong2024diversified} optimizes directional diversity and Qin\cite{qin2023infobatchlosslesstrainingspeed} achieves acceleration through selective pruning, both approaches have limitations in considering data learnability or relying on fixed thresholds. SLAP overcomes these limitations by combining stratified loss sampling with distance-based diversity control.

\textbf{Feature Selection.} Traditional embedding-based approaches in image processing \cite{sorscher2022beyond,zheng2022coverage,xia2022moderate} fail to capture dynamic training contributions and model changes. Recent gradient-based methods\cite{xia2024lessselectinginfluentialdata,wang2023farewell} show promise in dynamic feature capture and influence estimation. SLAP builds upon this direction, utilizing gradients for both coverage and diversity assessment.

\section{Method}

\vspace{-20pt}

\begin{figure*}[ht]
    \centering
    \includegraphics[width=1.0\textwidth]{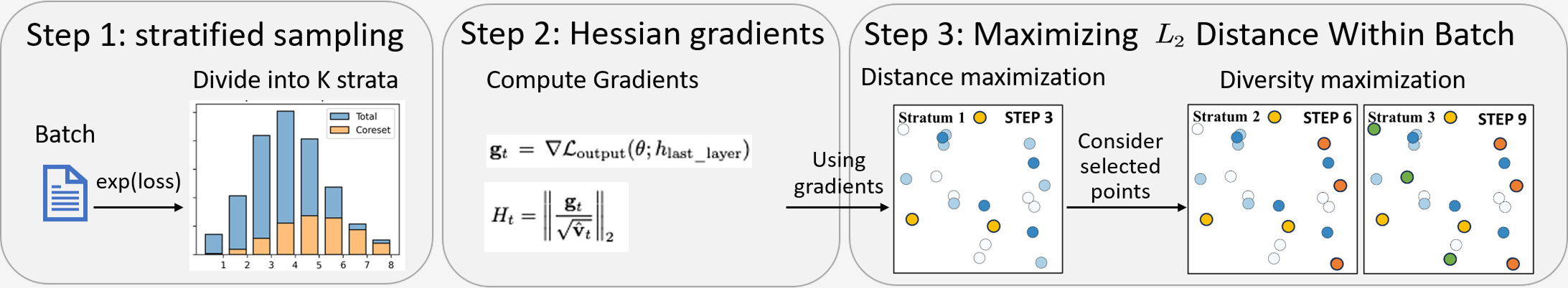}
    \vspace{-20pt}
    \caption{\textbf{The workflow of SLAP.} \textbf{Step 1:} We divide a batch of data into \( K \) strata based on loss. Then, we select \( |S| \) data according to the probability of normalized exp(loss) and calculate the number of data in each stratum. \textbf{Step 2:} We calculate the Hessian-approximated gradient \( H_t \) of the data as features. \textbf{Step 3: }For stratum 1, we randomly initialize a point. We calculate the $L_2$ distance to the first point and select the farthest point from the same stratum as the second point. We update the minimum distance from the remaining points to the selected point and repeatedly choose \( |S_i| \) samples. For strata 2 and 3, in order to select points for the new stratum, we need to consider the selected points from the previous strata. Finally, we will obtain a diversified subset that is relatively far from each other.}
    \label{fig:workflow}
\end{figure*}

\vspace{-10pt}

Here, we describe how SLAP adapts the gradient and stratified sampling to select samples that effectively induce a target capability. In \autoref{sec:3.1} we begin with a theoretical analysis of how to select a coreset that can approximately replace the full set through geometric analysis. Our analysis extends traditional coverage and diversity metrics to high-dimensional gradient spaces through rigorous mathematical derivation. 
Given the significant convergence and precision of Hessian when approaching the optimal solution, we utilize the Hessian-approximated gradient throughout the optimization process (\autoref{sec:3.2}).
In \autoref{sec:3.3}, we detail the incorporation of the SLAP framework, wherein we approximate the theory presented in \autoref{sec:3.1} through stratified sampling, complemented by dynamic on-policy batch selection, which facilitates O(n) computational efficiency.

\subsection{Geometric coverage of Coreset Selection}
\label{sec:3.1}
Ozan Sener\cite{sener2017active} explained coreset selection from the perspective of geometric spatial distribution, and proved that the assumption of zero training error holds when the model loss function satisfies Lipschitz continuity. The zero training error assumption means that for a model trained on a coreset, a bounded risk can be attained across the entire training dataset. Zheng\cite{zheng2022coverage} analyzed computer vision tasks based on the aforementioned theory. 

Below we provide an analysis for NLP tasks to explain the LLM model trained on coreset has a training bounded risk with radius $r$ covering the full set. We represent the full dataset of NLP task as \( S = \{(x_i, y_i)\}_{i=1}^{N} \), \( x_i = (x^1_i, x^2_i, \ldots, x^t_i) \) is the input token sequence and \( y_i \) is the prediction sequence \( y_i = (y^1_i, y^2_i, \ldots, y^t_i) \), where $y^t_i \in [\mathcal{C}]$ is the vocabulary label for token. A model trained on a coreset $S'$ has a training risk bounded with the covering radius $r$, if the loss function $l(x, y, h_S)$ is $\lambda_{l}$-Lipschitz continuous for all $y$ with bounded by $L$(Lipschitz constant), and the cross-entropy loss function $ l(x, y, h_S') = -\sum_{i\in\mathcal{S'}} y_i \log(\hat{y}_i)$ is $\lambda_{\eta}$-Lipschitz continuous. $h_S'$ is $r$ cover of $h_S$ and $|l(x, y; h_S) - l(x, y; h_S')| = 0$, $\forall (x, y) \in S'$. Then use Hoeffding's Bound and conclude that with probability at least $1-\gamma$: 

\vspace{-20pt}

{\small
\begin{align}
\left|\frac{1}{|S|}\sum_{i\in\mathcal{S}} l\left(x_i, y_i; h_{\mathcal{S}}\right) - \frac{1}{|S'|}\sum_{j\in\mathcal{S}} l\left(x_j, y_j; h_{\mathcal{S}^{\prime}}\right) \right|
\leq r\left(\lambda_{l} + \lambda_{\eta} L C\right) + \sqrt{\frac{L^2 \log\left(\frac{1}{\gamma}\right)}{2n}}
\end{align}
}

It means that given a full-set $S$ and a cover radius $r$ for the $S$, we can get a cover percentage $p$ coverage on the original distribution $P\mu$. The coreset is like a ball of radius $r$ that covers the full set and each sample is represented as a point in a high-dimensional space.



\subsection{Hessian-approximated gradient optimization}
\label{sec:3.2}
Common representations for sample typically rely on embeddings\cite{sorscher2022beyond,zheng2022coverage,xia2022moderate}, which primarily capture the intrinsic features of samples. However, this approach tends to overlook the importance of the model's influence during the training process. In contrast, we adopt gradient representations\cite{xia2024lessselectinginfluentialdata,wang2023farewell}, as gradients offer a dynamic measure of sample relevance, reflecting each sample's influence on model updates. And in high-dimensional space, the distances between gradient features facilitate a clearer differentiation of the relevance among samples.

In NLP tasks, we adopt a sequence-level gradient approach and leverage the sum of many token gradients from the last layer(\( \texttt{lm\_head} \)) of LLM to represent the entire sample sequence\cite{zheng2022coverage}. Summation is calculated because the feature weights of important tokens are preserved. The gradient at \( \texttt{lm\_head} \) captures highly abstracted features of the sample. The gradients from this layer are defined as follows:
\begin{align}
    \text{gradient}_{\text{lm\_head}}  = \nabla \mathcal{L}_{\text{output}}(\theta; h_{\text{last\_layer}}) \cdot h_{\text{last\_layer}}^T 
\end{align}
The term \( \nabla \mathcal{L}_{\text{output}}(\theta; h_{\text{last\_layer}}) \) signifies the gradient of the output layer with the model parameters \( \theta \). \( h_{\text{last\_layer}}^T \) is the last layer hidden states. 
In addition, the $L_2$ norm\cite{Hsieh_Ananthanarayanan_Bodik_Bahl_Philipose_Gibbons_Mutlu_2018,Babenko_Lempitsky_2015} is used for the gradient from the \( \texttt{lm\_head} \) layer.

Given the significant convergence and precision of Hessian when approaching the optimal solution, we substitute Hessian-approximated gradient optimization for the original gradient norm. 
Hessian-approximated gradients $H_{t}$ are derived from the norm of the gradient \( \mathbf{g}_t = \nabla \mathcal{L}_{\text{output}}(\theta; h_{\text{last\_layer}}) \) , adjusted by the second moment:
\begin{equation}
H_{t} = \left\| \frac{\mathbf{g}_t}{\sqrt{\hat{\mathbf{v}}_t}} \right\|_{2}
\end{equation}
where \( H_t \in \mathbb{R}^{D} \), \( D \) denotes the vocabulary size of the model. $\hat{\mathbf{v}}_t$ is the second moment estimation.

\subsection{SLAP: Optimizing On-Policy Batch Selection}
\label{sec:3.3}
In \autoref{sec:3.1}, an upper bound for the loss function of the coreset selection is an NP-Hard\cite{sener2017active,book_combinatorial_optimization} problem. We employ a stratified sampling method to effectively approximate a search across the entire range of the global set.
As illustrated in \autoref{fig:workflow}, SLAP initially computes the loss for each individual sample through forward propagation\cite{jiang2018mentornet,wei2020combating} and employs stratified sampling to select a representative subset from a batch samples \(B\) in each training step. Specifically, we select \(n_i\) samples from each stratum with the objective of maximizing the distance between the Hessian-approximated gradients of the selected samples.

\textbf{Distribution-aware stratified sampling based on loss probability. }
To align our sampling with the overall data distribution and account for both challenging and easy samples, we use a loss probability-based stratification approach. This method is detailed in ~\autoref{alg:1}.
We partition the batch into $K$ strata based on loss values. The width of each stratum is determined by the formula $(\text{max\_loss} - \text{min\_loss})/K$. Subsequently, we calculate the size of the selected sample set as $|S| = |B| \times \alpha$, where $\alpha$ denotes the pruning rate.
In our sampling process, we use \textbf{normalized exp(loss)} as the selection probability for each sample and we sample from the dataset without replacement to create the initial set. These samples are then categorized into $K$ strata, allowing us to tally the number of samples in each stratum.

\begin{figure}[h]
    \centering
    \includegraphics[width=0.9\linewidth]{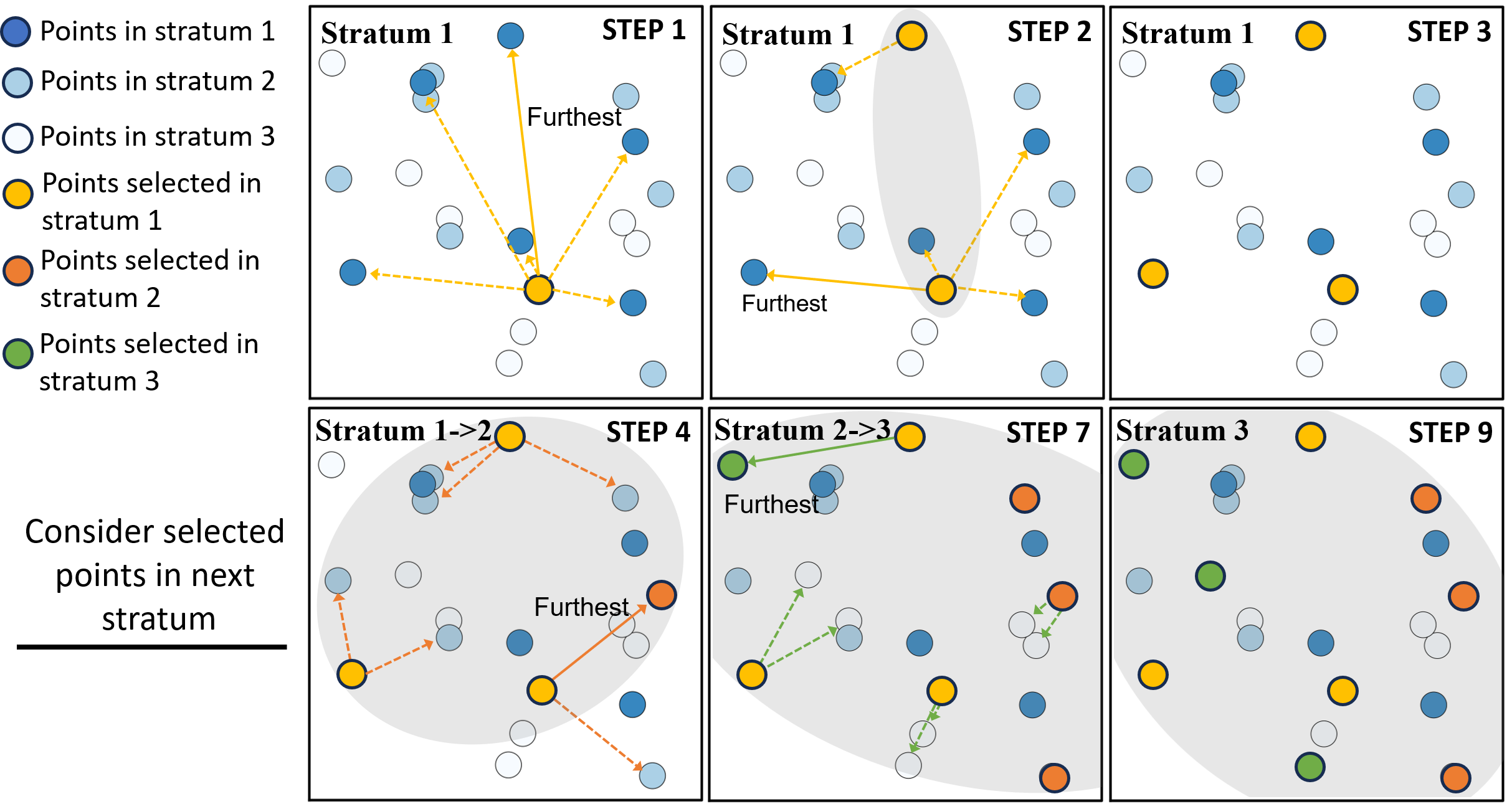}
    \caption{\textbf{Maximizing $L_2$ Distance Within the Batch.} Step 1: For stratum 1, randomly initialize a point and calculate the \( L_2 \) distance from the points in the same stratum to the first point. Step 2-3: Select the point that is farthest from the first point as the second point, then update the minimum distance from the remaining points to the selected points and iteratively choose \( |S_i| \) (e.g. 3) samples. Steps 4 and 7: For stratum 2, consider the selected points from the previous strata. Calculate the minimum distance from the points in stratum 2 to the selected points and take the maximum one as the first point. Then, iterate to complete the selection for stratum 2. Step 9: Finally, obtain a subset that is relatively far apart and diverse within the batch.}
    \label{fig:method}
\end{figure}

\textbf{Maximizing $L_2$ Hessian-approximated gradient distance within the batch. }
In each stratum, we aim to maximize the $L_2$ Hessian-approximated gradient distance among points to ensure greater separation within the batch as illustrated in ~\autoref{fig:method}. Specifically, for the number of points to be sampled in each stratum, indicated by $S_i$, we begin by randomly selecting one point. Subsequently, we compute the distances from the remaining points in the stratum to this selected point, referred to as $distance$. We then identify the point that is farthest away as the second point. We continue by calculating the $distance'$ from the remaining points to this second point, updating distance as $distance = min(distance, distance')$. This iterative process continues until we have selected the required number of points.


\begin{figure}[htbp]
    \centering
    \begin{minipage}{0.3\textwidth}
        \centering
        \includegraphics[width=\textwidth]{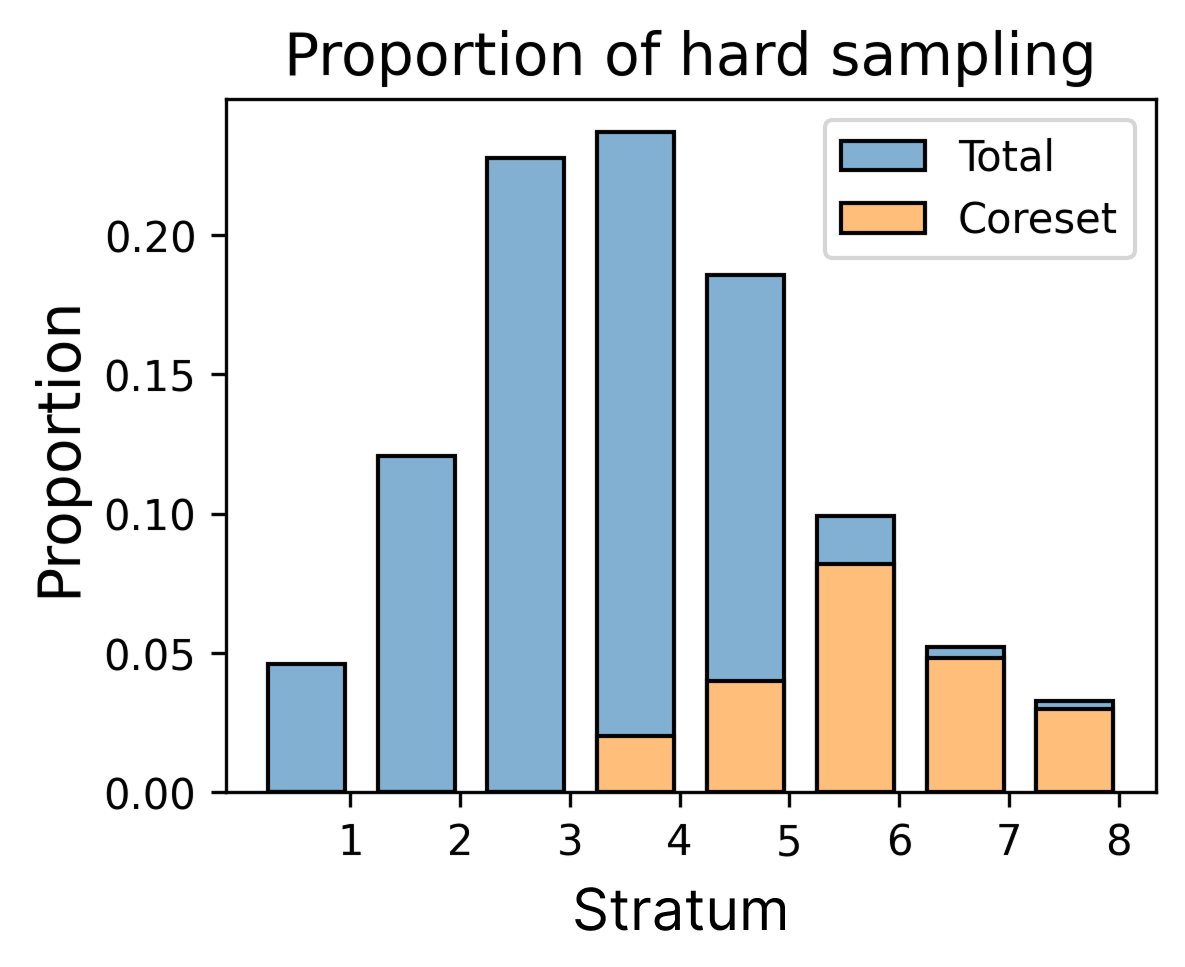}
    \end{minipage} 
    \hfill
    \begin{minipage}{0.3\textwidth}
        \centering
        \includegraphics[width=\textwidth]{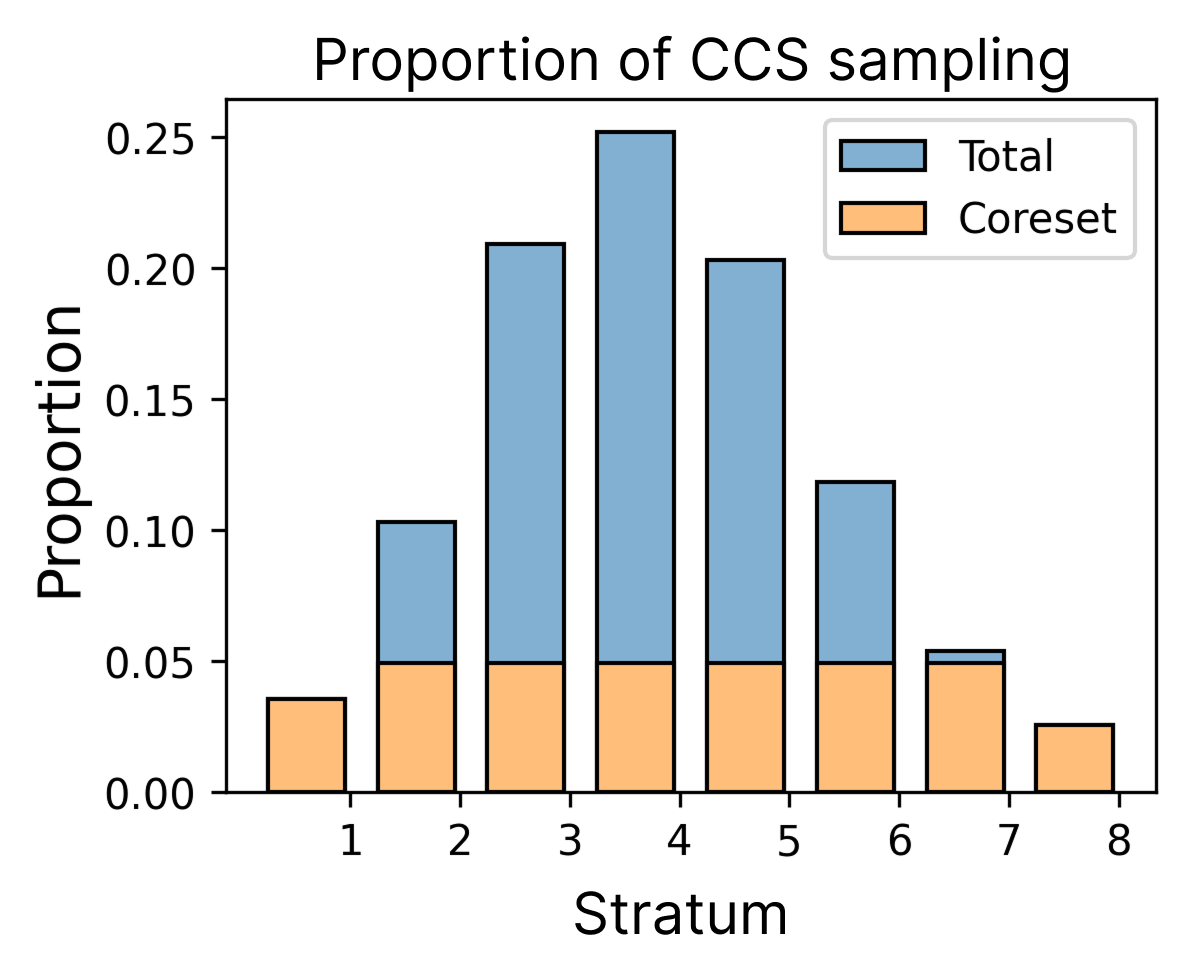}
    \end{minipage}
    \hfill
    \begin{minipage}{0.3\textwidth}
        \centering
        \includegraphics[width=\textwidth]{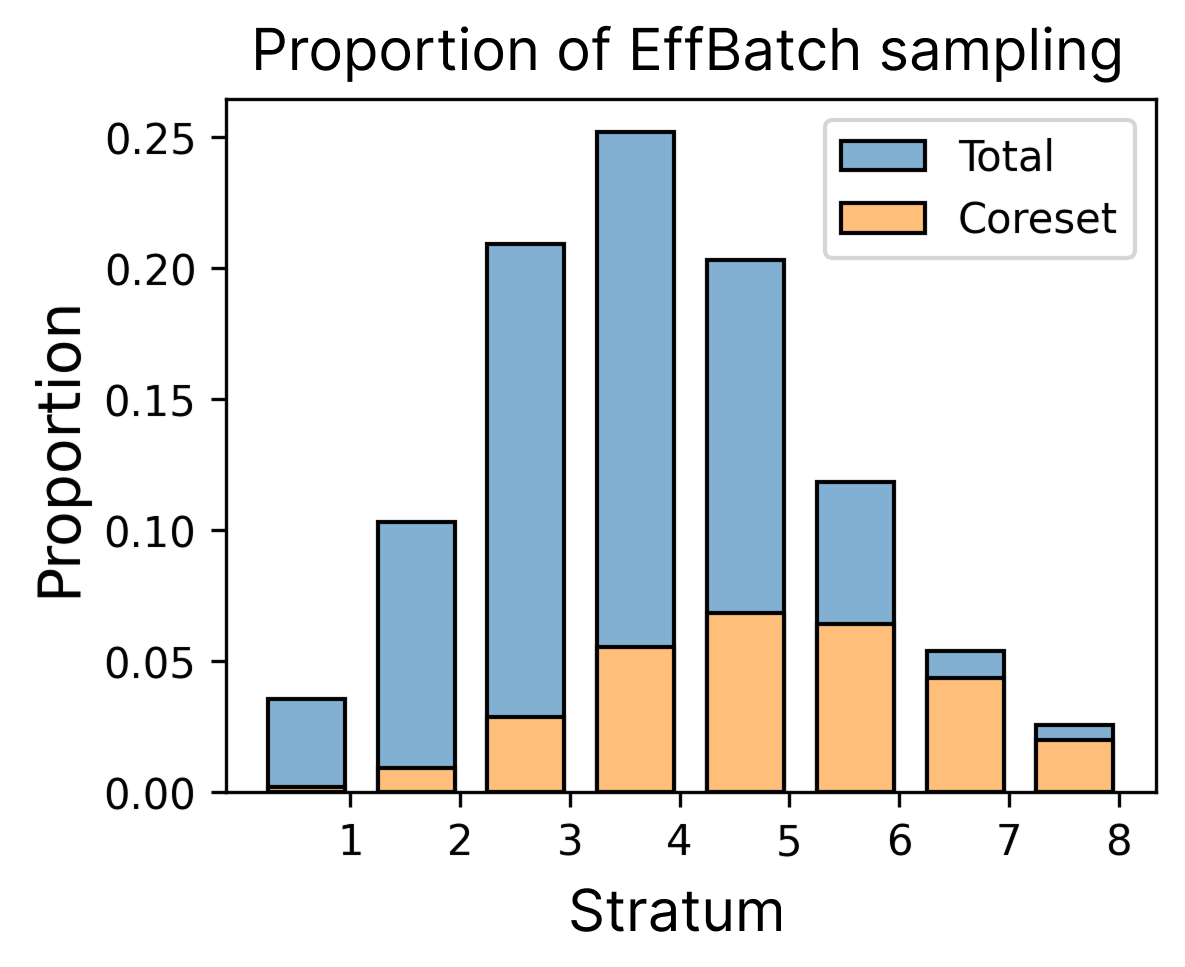}
    \end{minipage}
    \caption{The data distribution under hard sampling, CCS, and SLAP.}
    \label{fig:compare_optb_ccs_hard}
    \vspace{-10pt}
\end{figure}


We compare the coreset distribution of our data selection strategies with existing strategies further. As shown in \autoref{fig:compare_optb_ccs_hard}, hard sampling prioritizes the selection of samples starting from the most challenging (highest scoring) ones. Conversely, CCS emphasizes coverage, ensuring nearly equal representation of samples across different score ranges. SLAP integrates these approaches, balancing the consideration of both difficult and simple samples. This strategy prevents the model from experiencing excessive initial difficulty, thereby avoiding a significant increase in loss.

\begin{algorithm}[h]
    \caption{SLAP}
    \label{alg:1}
    \small
    \SetAlgoLined
    \KwIn{Batch $B$; current model $f_{\theta}$; pruning rate $\alpha$; the number of strata $K$; Hessian-approximated gradient $H_t$. }
    \KwOut{coreset $S$}
    \BlankLine
    loss $\leftarrow$ $f_{\theta}(B)$;\\ 
    coreset size $|S|$ = $\alpha$ * $|B|$; \\
    coreset $S$ $\leftarrow$ Sample $|S|$ samples without replacement according to the probability exp(loss); \\
    $B_0,B_1,...,B_{K-1}$ $\leftarrow$ Split loss in $|B|$ into $K$ ranges with an even range width; \\
    $|S_0|,|S_1|,...,|S_{K-1}|$ $\leftarrow$ Number of points selected in each stratum computed by $exp(loss)$; \\
    $S_{00} \leftarrow$ Randomly select the first point in $B_0$; \\
    \For{i in range(0, $K-1$) }{
        $distance \leftarrow \min(L_2$ Hessian gradient distances of $B_i$ to selected points $S) $;\\ 
        \For{j in range(0, $|S_i|$)} {
            $S_{ij} \leftarrow \arg\max(distance)$; \\
            $distance'$ $\leftarrow$ Calculate the distances of the remaining points to $S_{ij}$; \\
            $distance \leftarrow \min(distance, distance')$; \\
        }
    }
    \Return{\text{coreset} $S$};\
\end{algorithm}


\section{Experiment}
\subsection{Experimental Setup}
\textbf{Datasets.}
\setcounter{footnote}{0} 
We conduct our finetuning on the following instruction tuning datasets: (1) Net Literature Dialogue Dataset (NetLit); (2) llama3-Chinese-chat (LLaMaQA) \footnote{https://modelscope.cn/datasets/baicai003/Llama3-Chinese-dataset}; (3) WikiMatrix\footnote{https://github.com/facebookresearch/LASER/tree/main/tasks/WikiMatrix}\cite{schwenk2019wikimatrixmining135mparallel}.
NetLit consists of multi-turn dialogues related to web literature. Each entry is 1000 characters long, with a total of 100 million training samples. 
LLaMaQA is the dataset for the Chinese version of LLaMa3, which includes system instructions, queries, and GPT-4 responses, amounting to 1.69 million training samples. 
WikiMatrix extracts parallel sentences in all possible language pairs from the content of Wikipedia. We use 10 language pairs with 28 million training samples.

\begin{figure}[ht]
    \centering
    \begin{minipage}[t]{0.44\textwidth}
        \centering
        \includegraphics[width=\linewidth]{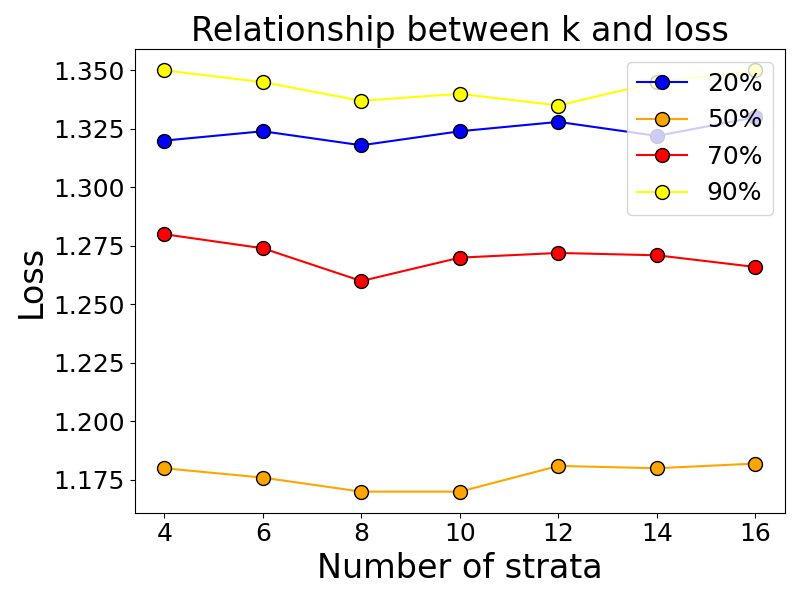} 
        \caption{Evaluation with different k on NetLit using ChatGLM3 model.}
        \label{fig:open}
    \end{minipage}
    \hspace{0.05\textwidth} 
    \begin{minipage}[t]{0.45\textwidth}
        \centering
        \includegraphics[width=\linewidth]{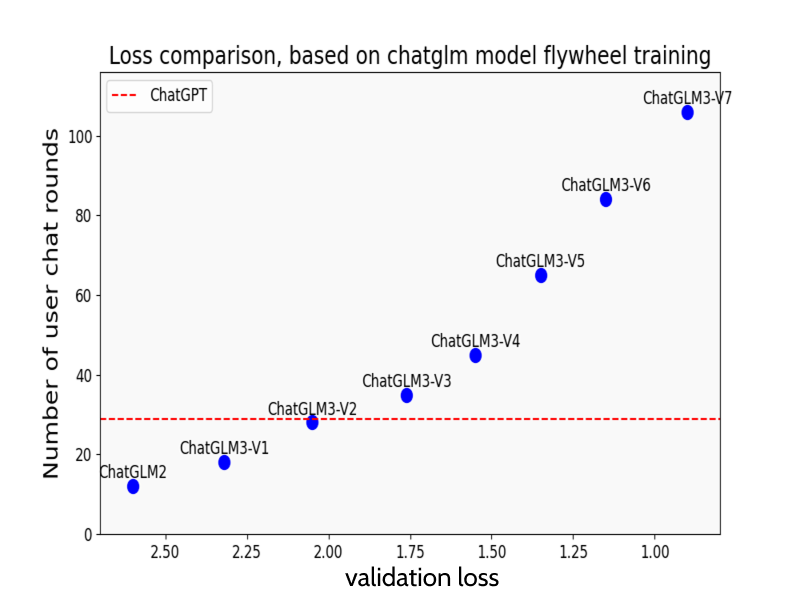} 
        \caption{Loss comparison based on ChatGLM model flywheel training.}
        \label{fig:loss_and_round}
    \end{minipage}
\end{figure}

\textbf{Baselines.} 
We compare our \textbf{SLAP} with various baseline methods, including random selection, online hard, CCS, and InfoBatch. 
\textbf{Random selection} randomly select data from training datasets. 
\textbf{Online hard} starts by selecting data points with the highest loss and continues downwards until enough data points are chosen.
\textbf{CCS}\cite{zheng2022coverage} stratifies the data based on loss and randomly selects a fixed number of data points from each stratum. It utilizes important scores derived from the computer vision domain;  We did the same experimental analysis for the number of strata $k$ (\autoref{fig:open}). Since the number of strata is not a sensitive hyperparameter. The default number of k for CCS is 50, but we use 8 instead.
\textbf{InfoBatch}\cite{qin2023infobatchlosslesstrainingspeed} randomly removes samples with low information content based on the loss distribution, then adjusts the gradients of the remaining samples to match the original gradients. The threshold is set to the average loss. Nevertheless, this approach proves inadequate when dealing with high pruning rates. To address this limitation, we increase the threshold appropriately for higher pruning rates.

\textbf{Implementation details.}
We train with two base models: 
Meta-LLama-3-8B-Instruct\cite{llama3modelcard} and ChatGLM-3-6B\cite{glm2024chatglm}. The hyper-parameters we used are as follows. We fine-tune the base model for 1 epoch with AdamW using a cosine learning rate scheduling strategy, $\beta_1=0.9$, $\beta_2=0.999$, $\epsilon=1e-8$. We set the initial learning rate is set to $1e-5$. The batch size is set to 320, the context window's maximum length is 2048 tokens, and longer samples are trimmed to fit in. 
We use 16 $\times$ A800 80G GPUs for training.

\textbf{Evaluation Metrics.}
\label{EM}
Through extensive empirical analysis spanning ten consecutive A/B experiments with a substantial user base (n > 200,000), we discovered a statistically significant negative correlation between model loss and user engagement metrics, specifically conversation turn length. As illustrated in \autoref{fig:loss_and_round}, lower loss values consistently correspond to higher user retention rates, manifesting as increased conversation duration. This robust correlation, validated with confidence less than 0.01 in all experimental iterations, provides a strong empirical justification for adopting \textbf{validation loss} as our primary evaluation metric.
Furthermore, considering the objectivity of the experiments, common metrics are used for evaluation, such as GPT4 evaluation, human expert evaluation\cite{shao2023character,wang2023rolellm}, BLUE, Rlouge\cite{papineni2002bleu,lin2004rouge}, etc.


\vspace{-20pt}

\subsection{Main results}
\begin{figure*}[h]
    \centering
    \begin{minipage}{0.32\linewidth}
        \centering
        \includegraphics[width=\linewidth]{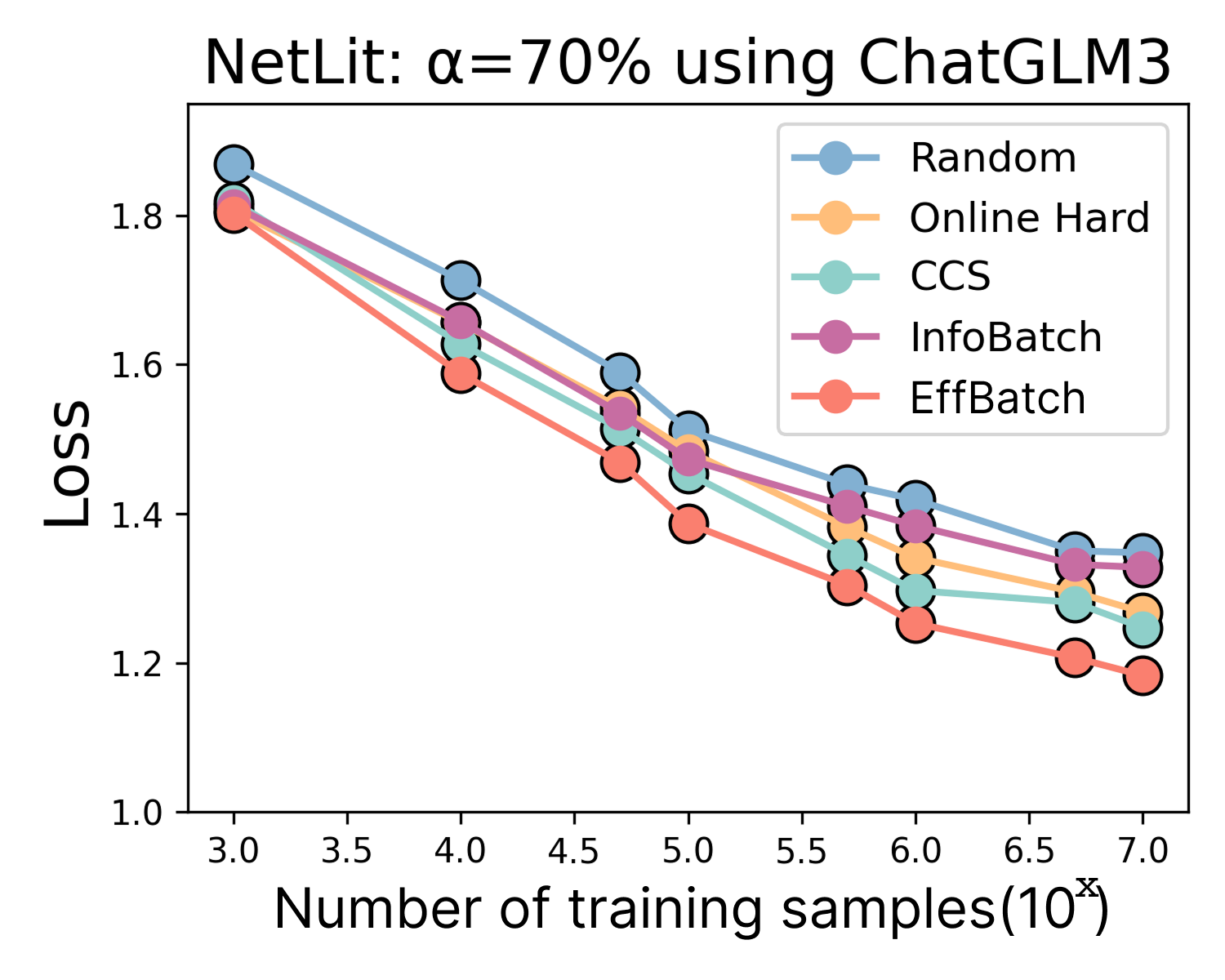}
    \end{minipage}
    \hfill
    \begin{minipage}{0.32\linewidth}
        \centering
        \includegraphics[width=\linewidth]{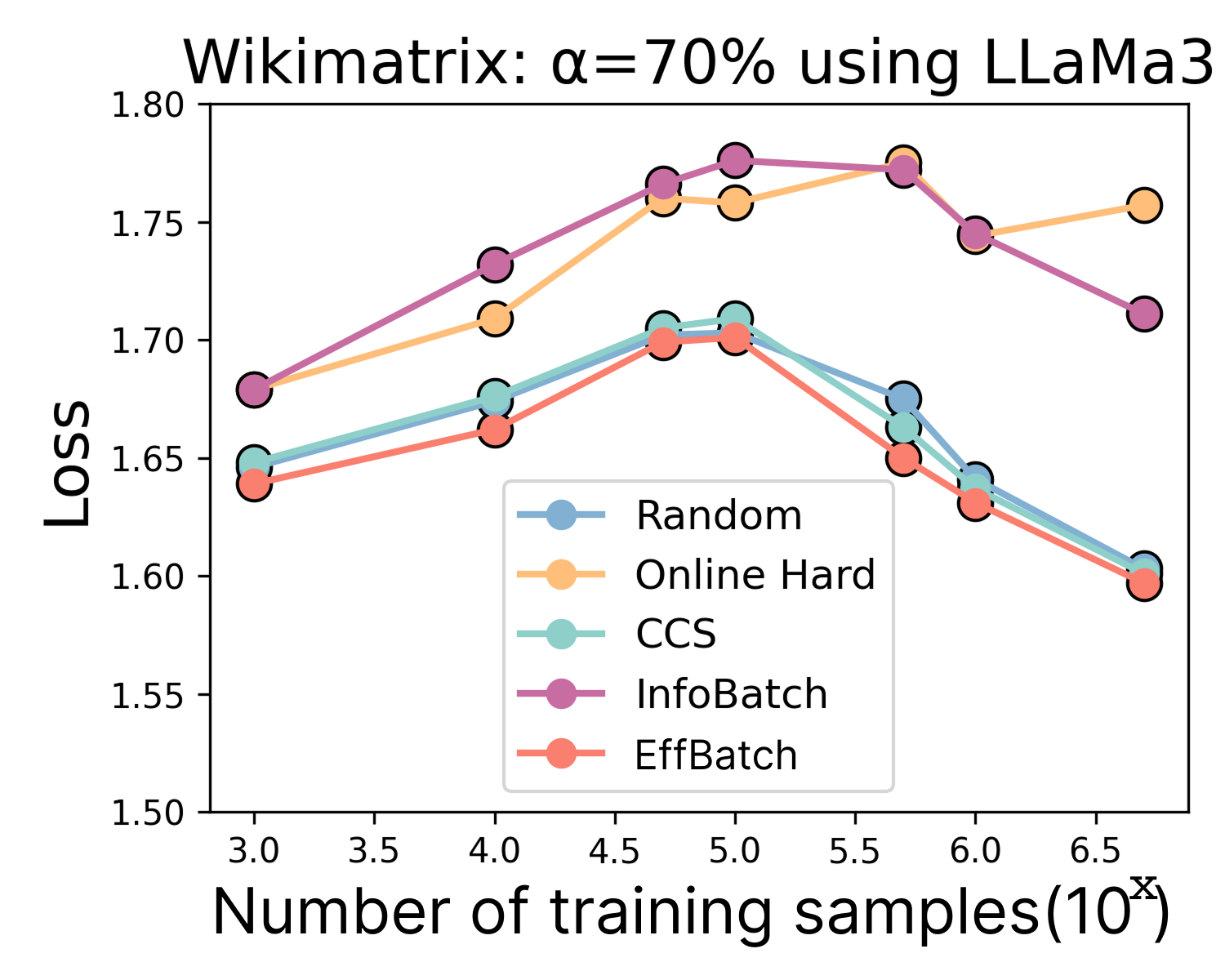}
    \end{minipage}
    \hfill
    \begin{minipage}{0.32\linewidth}
        \centering
        \includegraphics[width=\linewidth]{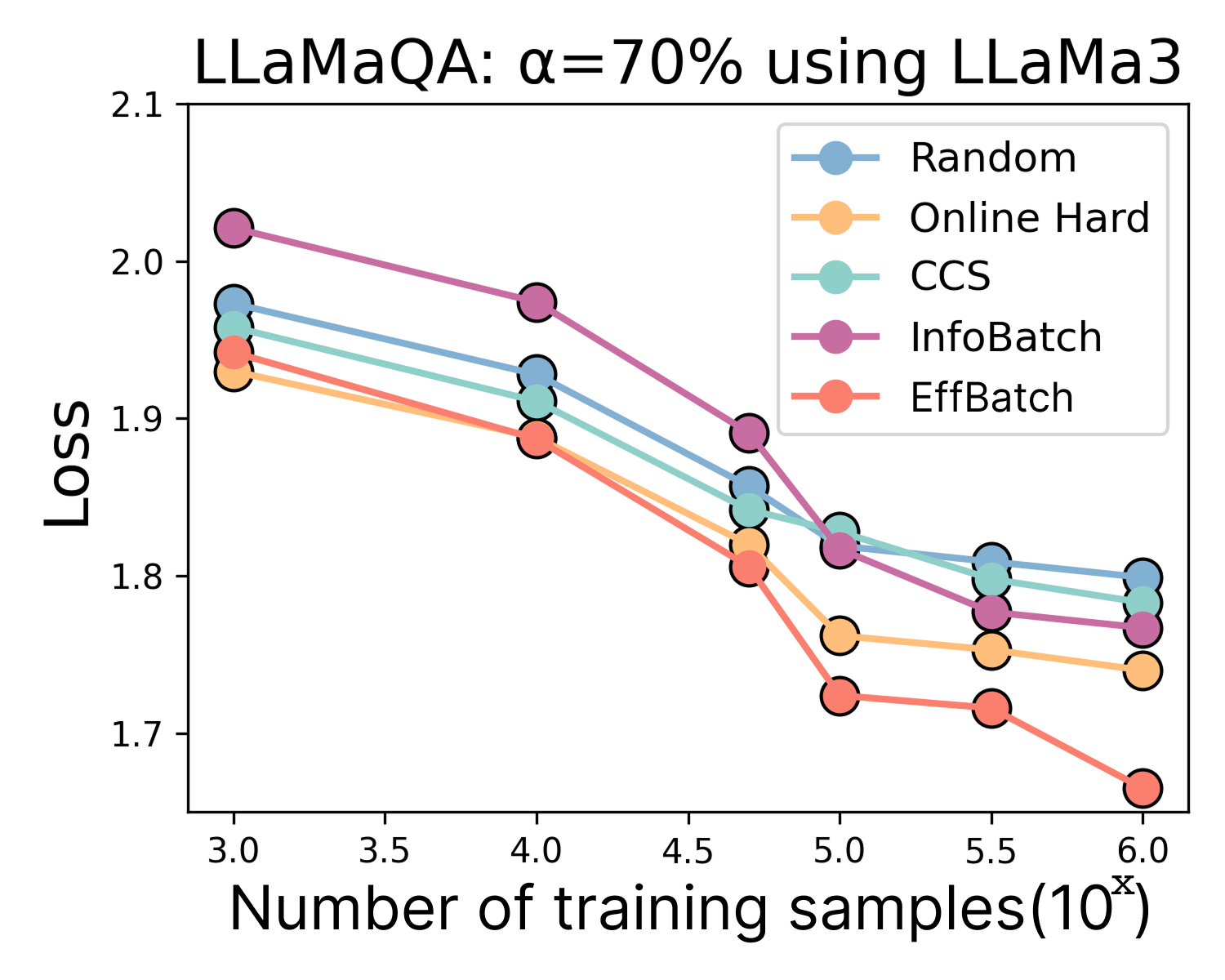}
    \end{minipage}    
    \caption{Evaluation on different datasets using ChatGLM3 model with pruning rate 70\%.}
    \label{fig:s2}
\end{figure*}

\textbf{Performance under different datasets.}
Based on the comparisons in ~\autoref{fig:s2}, we have the following observations: 
(1) SLAP exhibits superior performance across a variety of datasets. In particular, both the NetLit dataset and the LLaMaQA dataset reveal that SLAP consistently achieves a significantly low loss than other methods. 
(2) SLAP strikes a balance between diversity coverage and task difficulty, thereby achieving enhanced performance across all three downstream tasks. The performance of CCS and InfoBatch varies significantly across different datasets, presenting a contrasting trend. CCS demonstrates superior efficacy compared to InfoBatch in both the NetLit and WikiMatrix datasets, suggesting that the aspect of diversity coverage is particularly crucial within the realms of web literature and translation tasks. Conversely, the LLaMaQA dataset places a greater emphasis on the dimension of task difficulty. 

\vspace{-20pt}

\vspace{-20pt}

\begin{figure}[ht]
    \centering
    \begin{minipage}{0.33\linewidth}
        \centering
        \includegraphics[width=\linewidth]{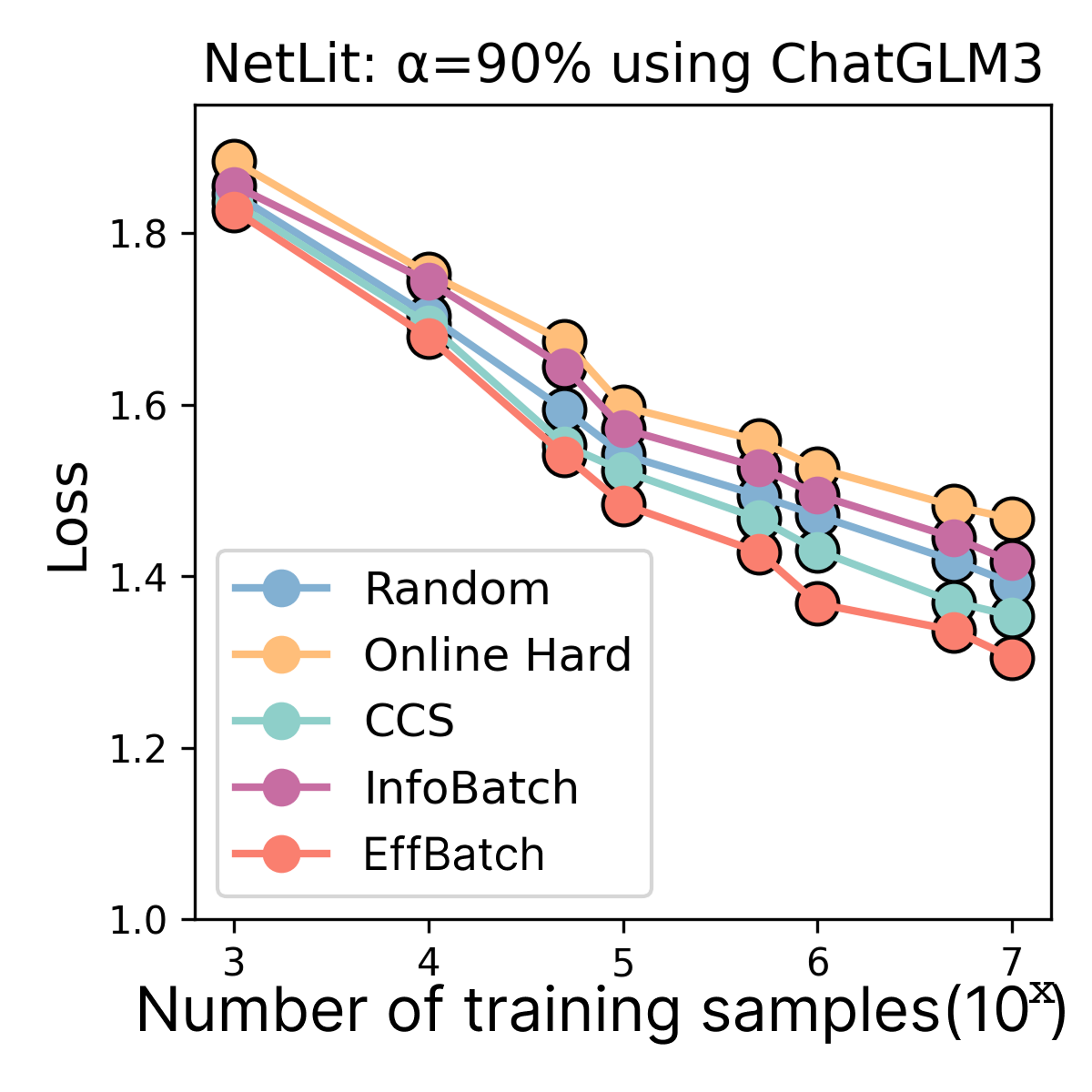}
    \end{minipage}
    \hspace{0.015\linewidth}    
    \begin{minipage}{0.33\linewidth}
        \centering
        \includegraphics[width=\linewidth]{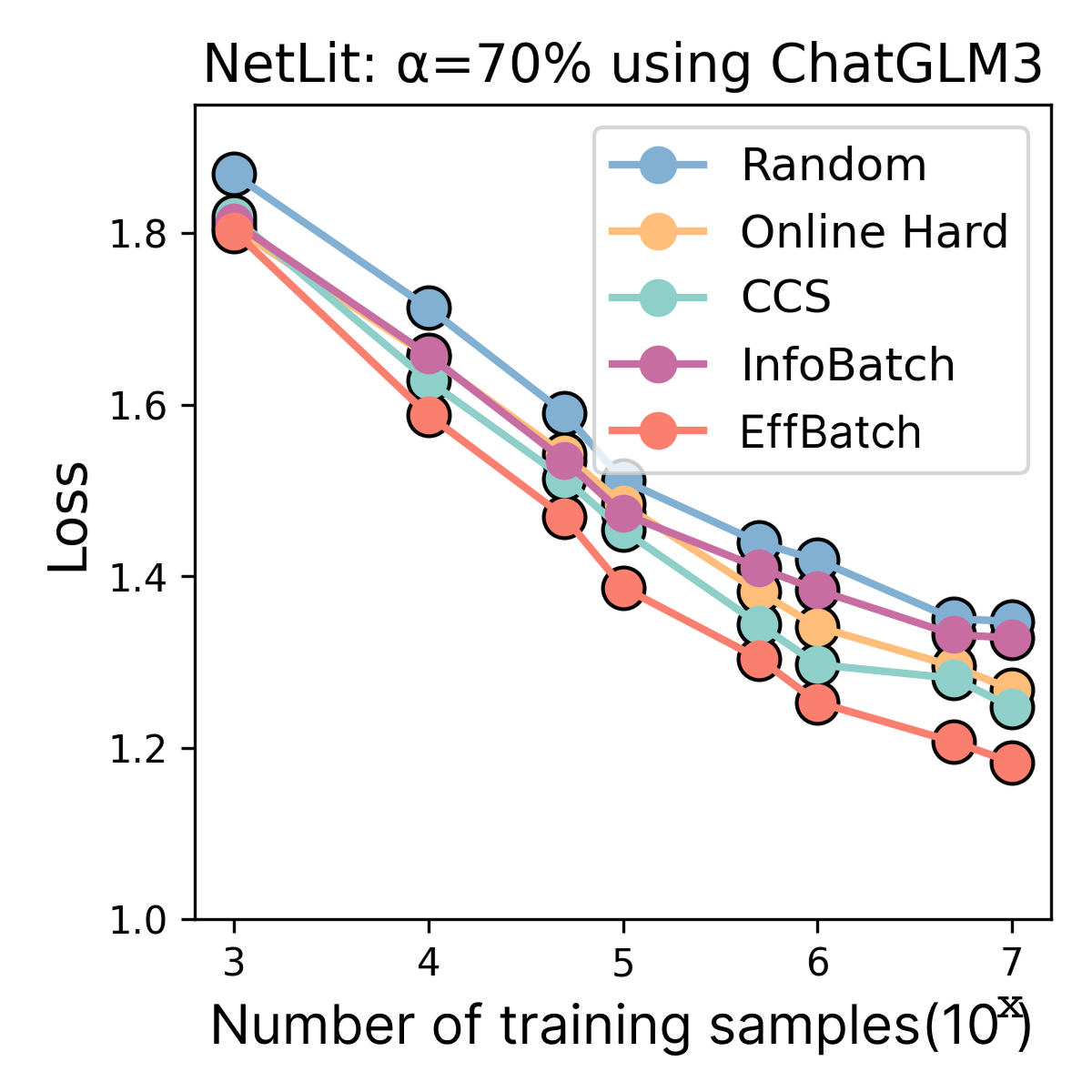}
    \end{minipage}
    \vfill  
    \begin{minipage}{0.33\linewidth}
        \centering
        \includegraphics[width=\linewidth]{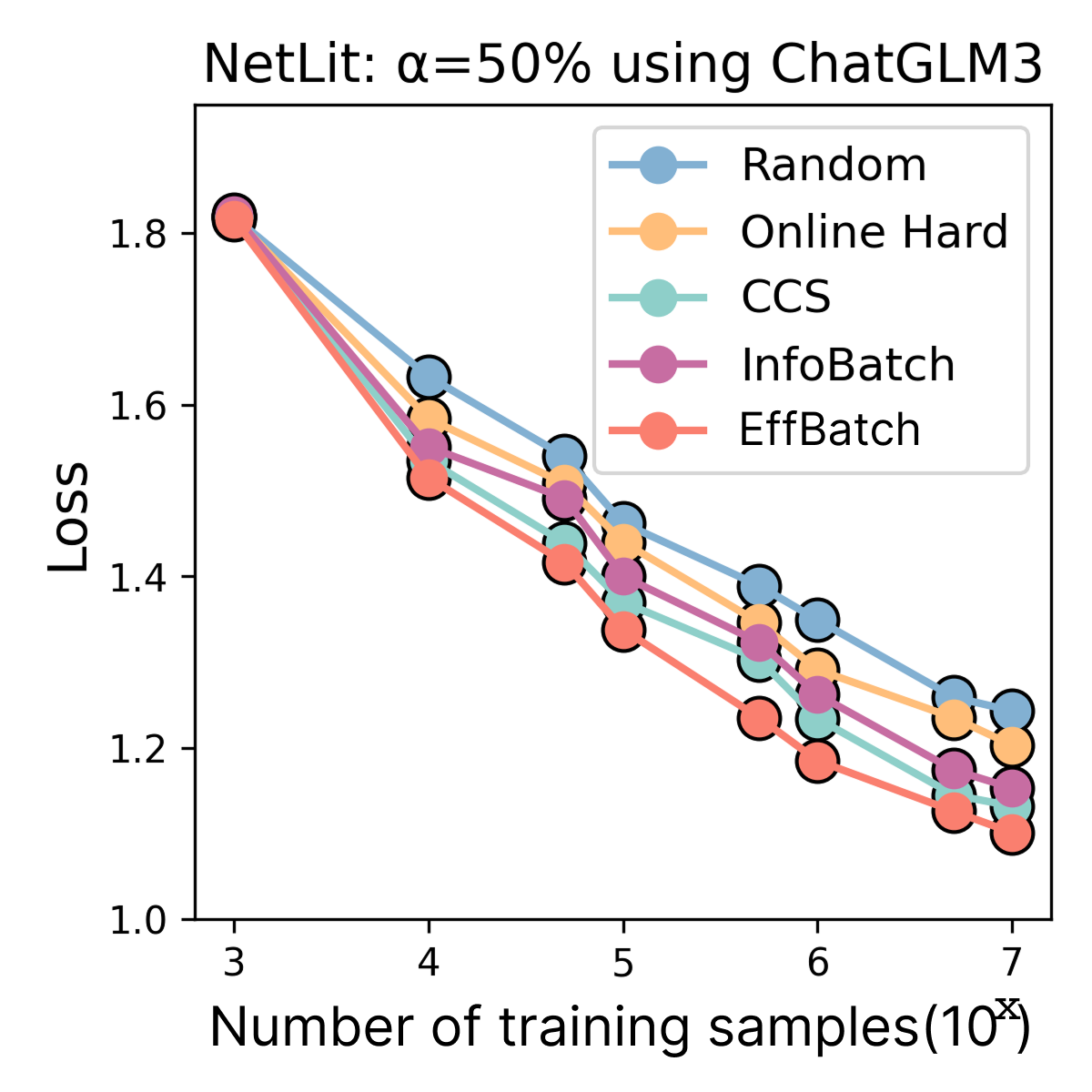}
    \end{minipage}
    \hspace{0.015\linewidth}    
    \begin{minipage}{0.33\linewidth}
        \centering
        \includegraphics[width=\linewidth]{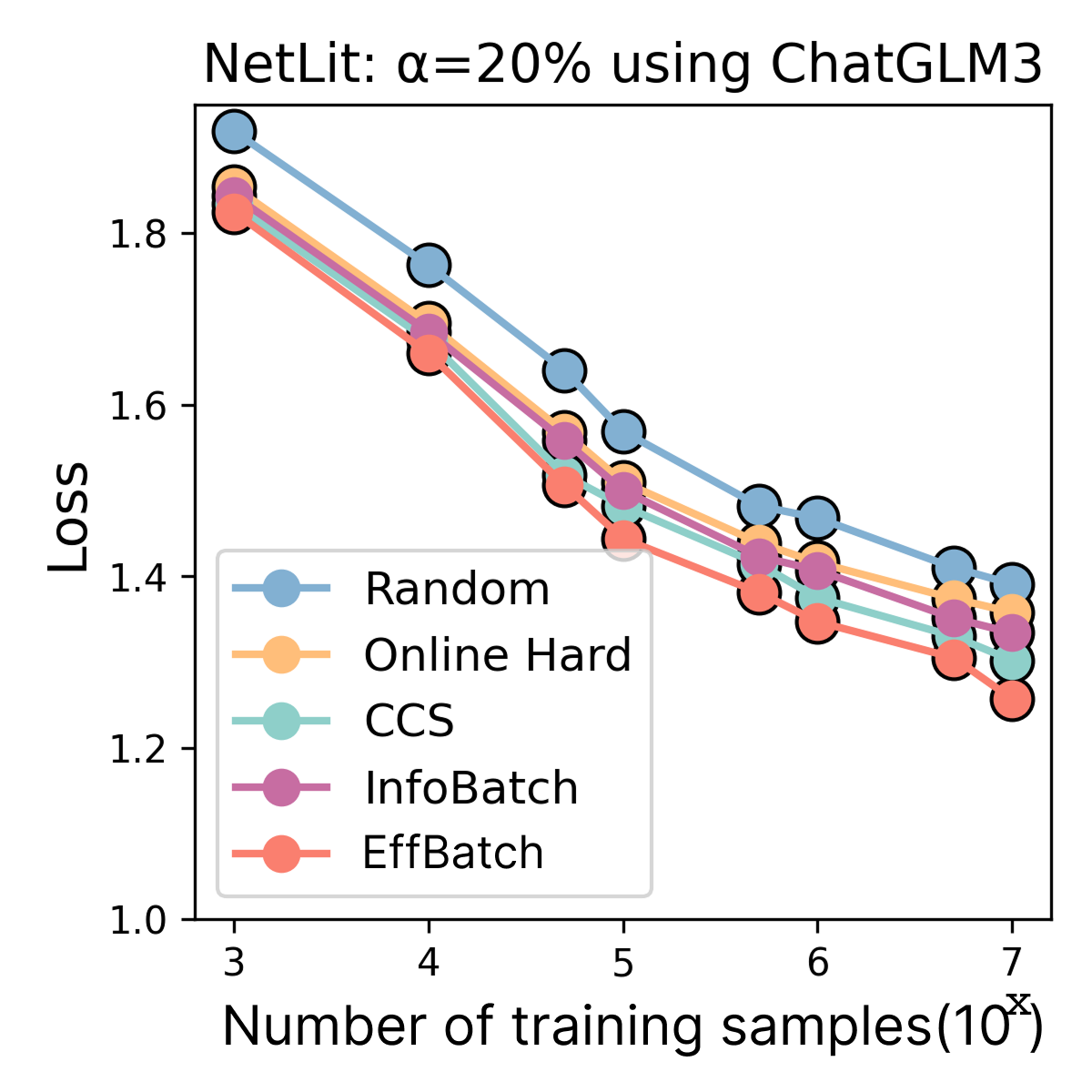}
    \end{minipage}
    \caption{Evaluation with different pruning rates on NetLit using ChatGLM3 model}
    \label{fig:3}
\end{figure}

\textbf{Performance under different pruning rates.} In ~\autoref{fig:3}, SLAP consistently demonstrates the lowest loss across various pruning rates. At $\alpha=50\%$ pruning rate, all methods achieved the lowest loss. This is primarily because both hard data and easy data have been retained to a certain extent. Notably, the losses at $\alpha=90\%$ and $\alpha=20\%$ are remarkably similar, likely due to the high redundancy in the web text domain. Therefore, we can train the model with less data to reduce computational costs in reality. 

\begin{figure}[ht]
    \centering
    \begin{subfigure}{0.4\textwidth}
        \centering
        \includegraphics[width=\textwidth]{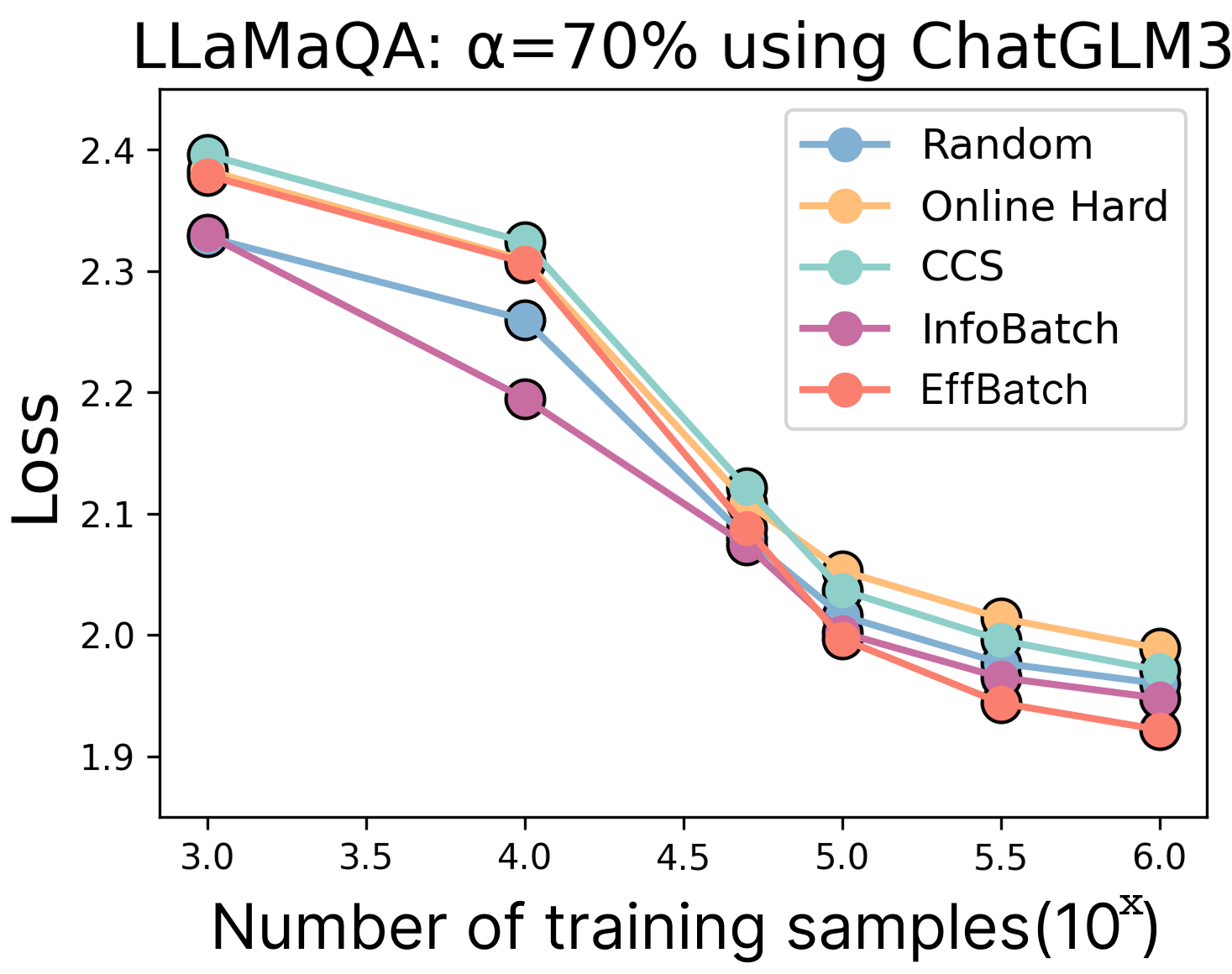}
    \end{subfigure} 
    \hspace{0.045\textwidth} 
    \begin{subfigure}{0.4\textwidth}
        \centering
        \includegraphics[width=\textwidth]{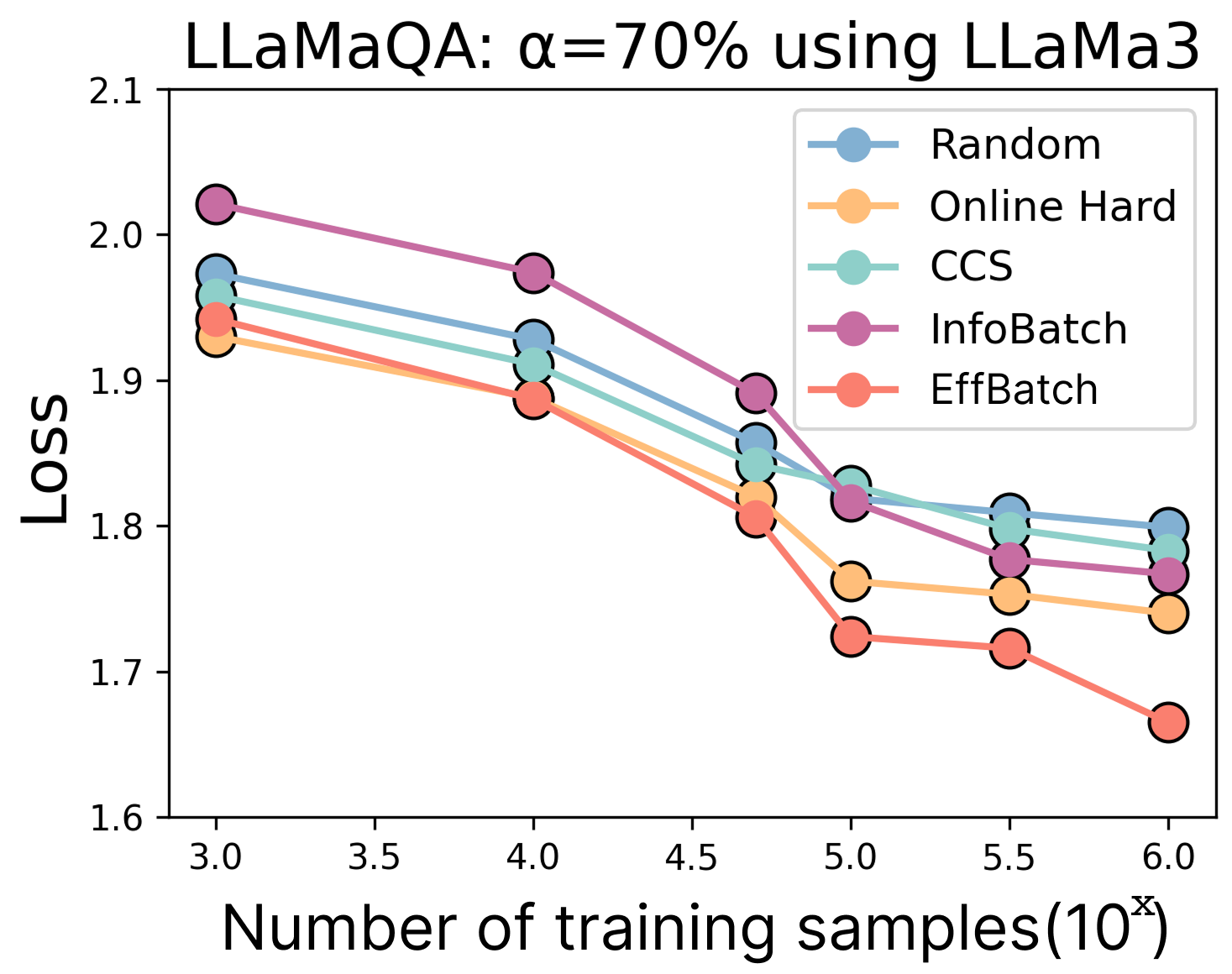}
    \end{subfigure} 
    \caption{Evaluation on LLaMaQA using ChatGLM3 and LLaMa3 with pruning rate 70\%.} 
    \label{fig:5}
\end{figure}

\textbf{Performance between different models.}
In ~\autoref{fig:5}, we find that SLAP exhibits superior stability across various models, consistently achieving the lowest loss values in both cases examined. In contrast, InfoBatch and CCS demonstrate varying degrees of instability across the same models. Specifically, SLAP is particularly effective for the ChatGLM3 model when applied to large datasets, while InfoBatch is more advantageous for smaller datasets. Conversely, the LLaMa3 model shows a stronger alignment with the Online Hard and SLAP methodologies.


\begin{figure}[ht]
    \centering
    \begin{subfigure}{0.45\textwidth}
        \centering
        \includegraphics[width=\textwidth]{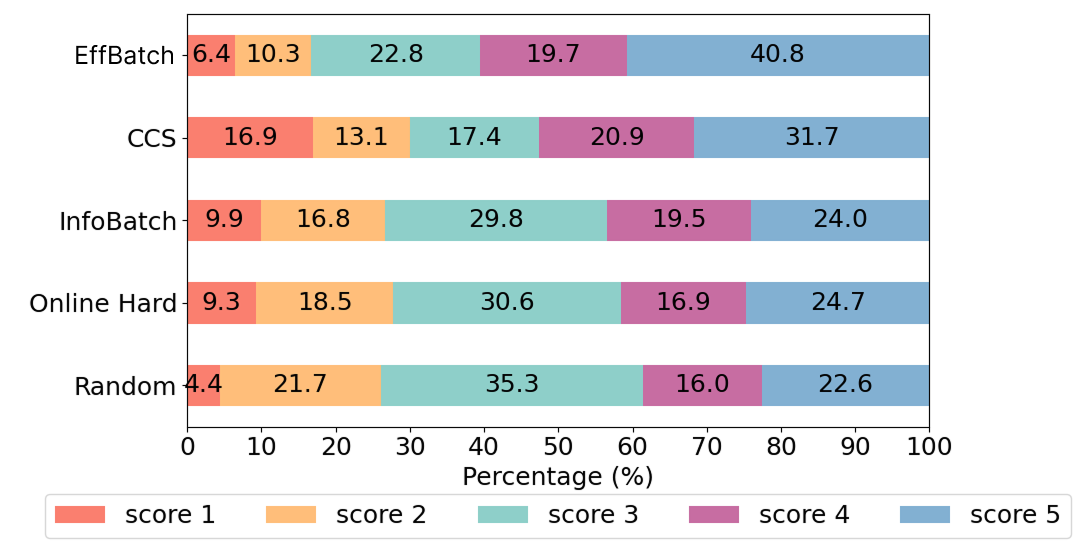}
        \caption{GPT-4 Evaluation.}
    \end{subfigure} 
    \begin{subfigure}{0.45\textwidth}
        \centering
        \includegraphics[width=\textwidth]{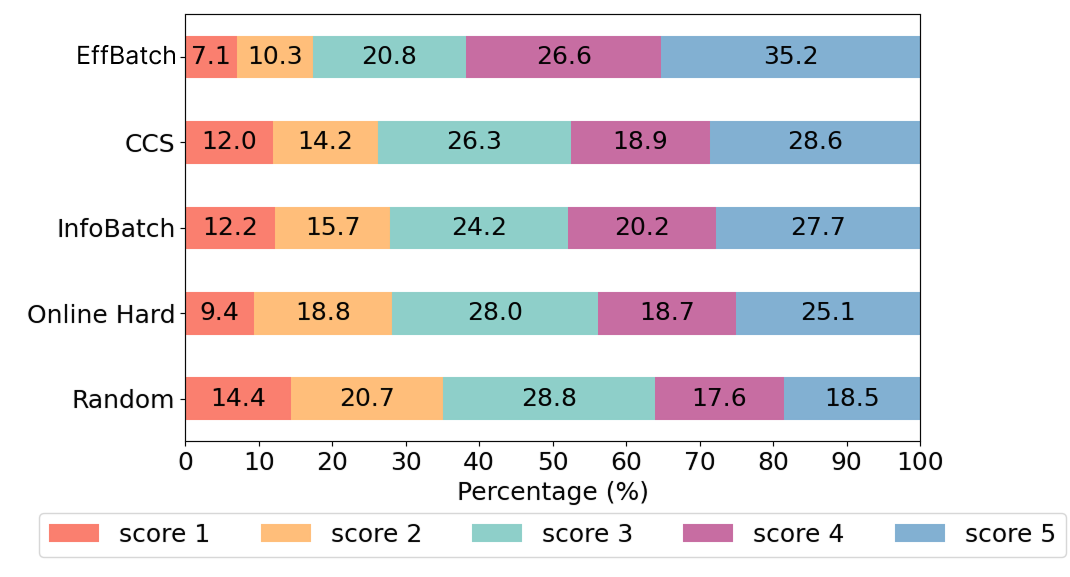}
        \caption{Human Evaluation.}
    \end{subfigure} 
    \caption{GPT-4 and human evaluation scores for LLM generated responses on NetLit dataset.} 
    \label{fig:gpt4_eva}
\end{figure}

\textbf{GPT-4 Evaluation.} 
For our net literature dataset NetLit, we are more concerned with the performance of LLMs to generate responses under various pruning methods, similar to the capabilities of LLMs in role-playing scenarios\cite{shao2023character,wang2023rolellm}. We evaluate the performance on personality and speaking style as the character's primary feature. 
We instruct GPT-4 to score the generated responses, and the Chain of Thought\cite{wei2022chain} process allows us to evaluate the performance of different pruning methods effectively. We refer to the prompt template\cite{shao2023character}. GPT-4 scores each response from 1 to 5, with 5 indicating strong alignment with the character's personality and speaking style, and 1 indicating poor alignment. In the evaluation results from GPT-4, both CCS and InfoBatch, along with our method, achieved commendable outcomes. Such as in ~\autoref{fig:gpt4_eva}(a), SLAP has the highest percentage of high-score samples, reaching 60.5\%. In comparison, the CCS method achieves 52.6\%, while the InfoBatch method stands at 43.5\%. Additionally, SLAP produces fewer low-score examples than the other methods. 

\textbf{Human Evaluation.} 
Furthermore, human judgment is still the most thorough and realistic assessment of whether the generated response is character-aligned. Some poor GPT-4 annotation cases are discovered during our task. As illustrated in ~\autoref{fig:gpt4_eva}(b), we invite annotators to rectify the scoring results of GPT-4 for each test data, leading to human evaluation results. Due to space limitations, detailed prompts for response generation and scores on other datasets are omitted here.

\textbf{Reference-based Metrics.} We employ further industry-recognized metrics Bleu-4\cite{papineni2002bleu}, Rouge-1, Rouge-2, and Rouge-L\cite{lin2004rouge} to validate the performance on different datasets. Consistent with previous experiments, we utilize the LLaMa3 model for training and validation on WikiMatrix and LLaMaQA dataset, while the ChatGLM3 model employed for NetLit datasets. All experiments are conducted at a pruning rate of 70\%. As shown in ~\autoref{tab:d1d1}, the results demonstrate that our SLAP approach significantly outperforms its approaches, highlighting its versatility and efficacy across a range of tasks and models.


\begin{table*}[htbp]
\centering
\caption{Quantitative comparison on NetLit | WikiMatrix | LLaMaQA dataset.}
    {\small
        \begin{tabular}{|c|c|c|c|c|}
        \hline
          & \textbf{Bleu-4} & \textbf{Rouge-1} & \textbf{Rouge-2} & \textbf{Rouge-L} \\
        \hline
        Random & 13.40|31.52|22.09 & 34.82|44.57|40.92 & 18.10|21.69|21.79 & 30.03|36.56|35.38 \\
        Online Hard & 10.43|24.46|27.04 & 28.10|39.62|43.52 & 11.86|13.84|22.74 & 24.11|30.22|38.77 \\
        InfoBatch & 13.16|26.76|25.31 & \textbf{35.30}|42.67|40.73 & 18.05|19.45|22.62 & 30.22|34.17|37.88 \\
        CCS & 13.03|\textbf{32.94}|26.17 & 34.60|46.17|42.57 & 17.71|\textbf{22.84}| 22.33 & 30.05|38.24|37.31 \\
        SLAP & \textbf{13.73}|32.24|\textbf{27.11} & 35.02|\textbf{47.51}|\textbf{44.07} & \textbf{18.25}|22.56|\textbf{22.81} & \textbf{30.53}|\textbf{38.97}|\textbf{39.52} \\
        \hline
        \end{tabular}
    }
\label{tab:d1d1}
\end{table*}


\subsection{Efficient Performance}
\begin{figure}[ht]
    \centering
    \begin{subfigure}{0.4\textwidth}
        \centering
        \includegraphics[width=\textwidth]{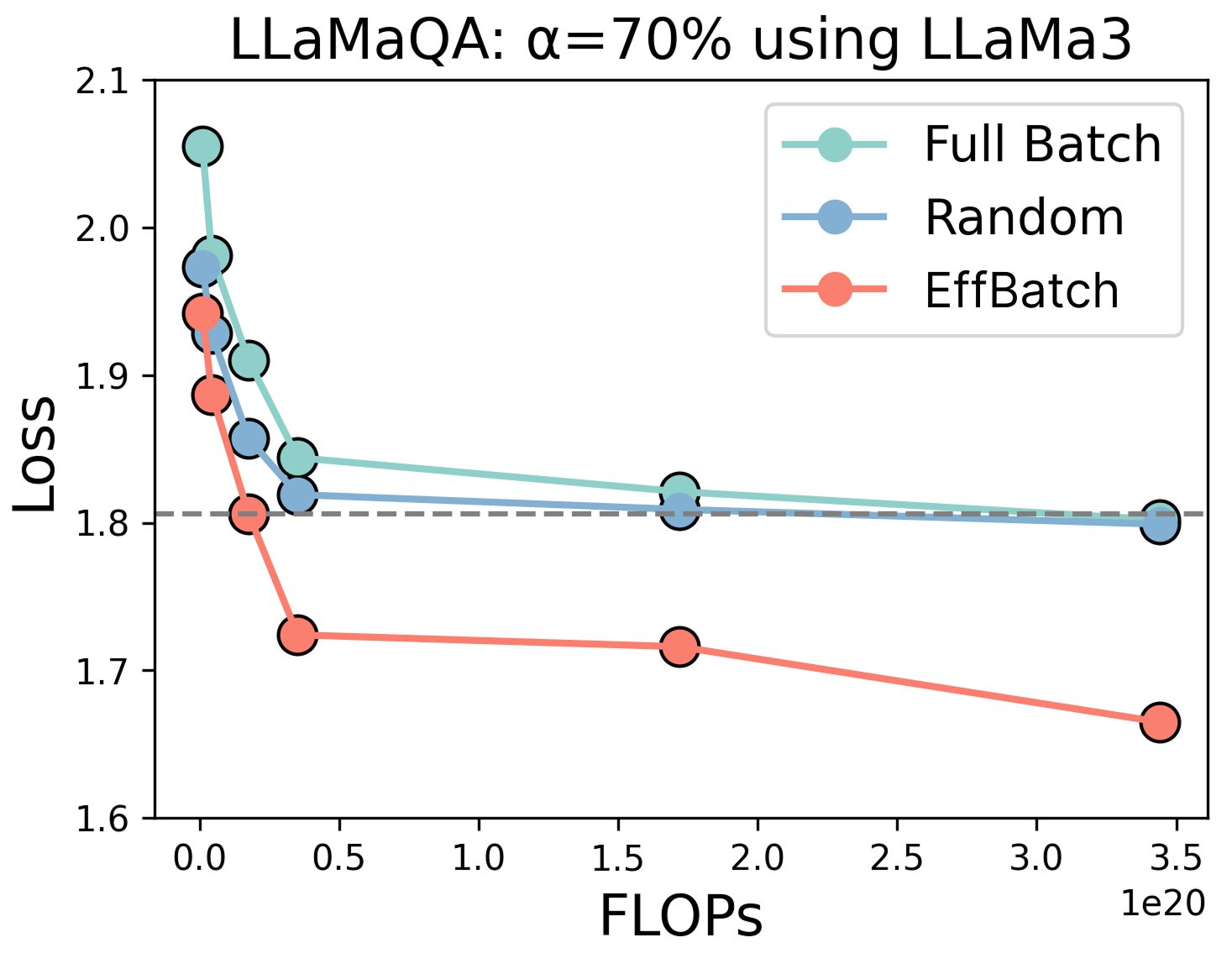}
    \end{subfigure} 
    \hspace{0.045\textwidth} 
    \begin{subfigure}{0.4\textwidth}
        \centering
        \includegraphics[width=\textwidth]{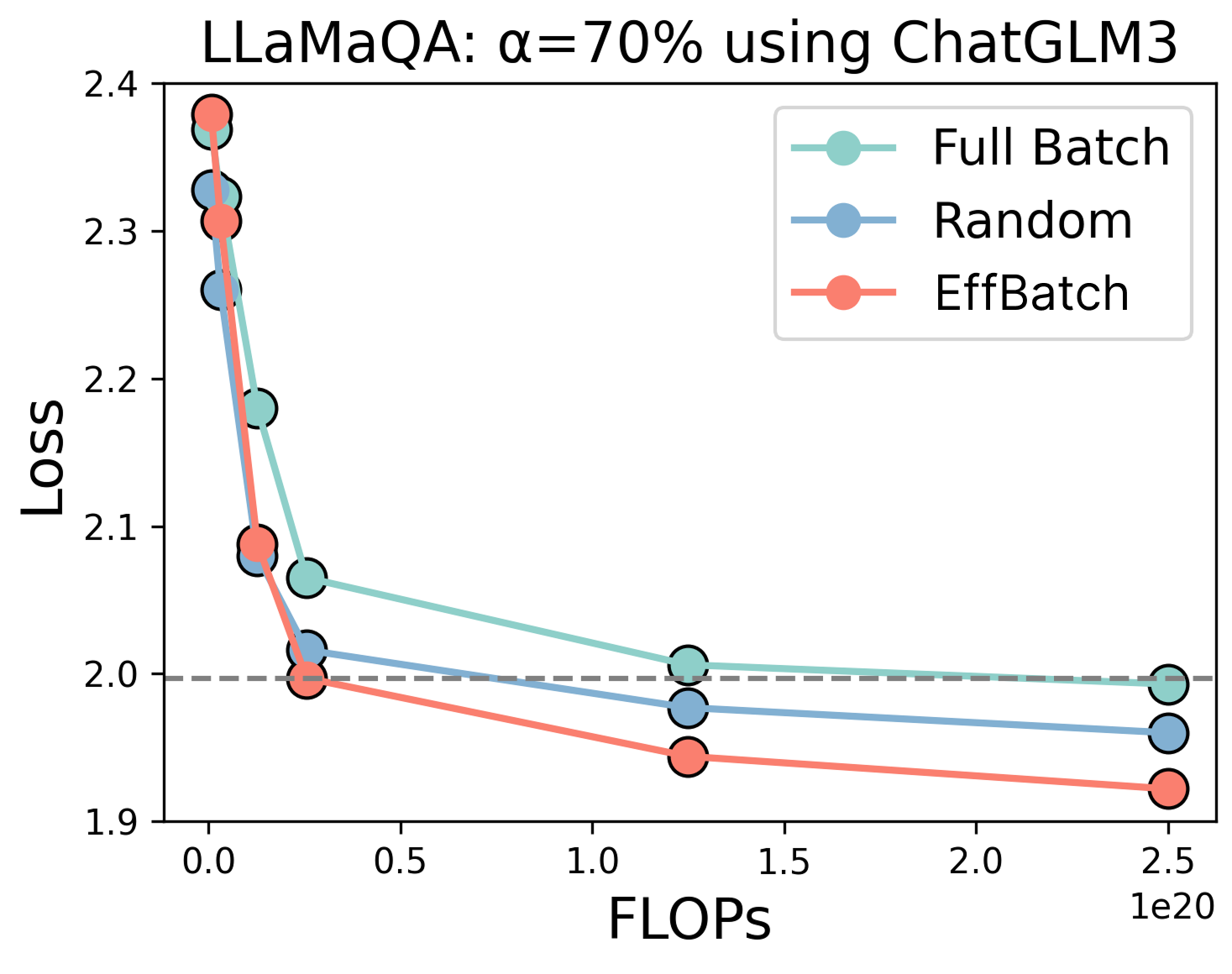}
    \end{subfigure} 
    \caption{Comparison of FLOPs for Pruning and Full data.} 
    \label{fig:d2}
\end{figure}

To quantify the computational savings achieved by our method, we compare the FLOPs (Floating Point Operations) required for both full batch data and pruned batch data using SLAP. For full batch processing, the computation is represented by the formula:
\begin{equation}
\begin{array}{l}
\text{Full-Batch FLOPs} = B \times (L_i + L_o) \times F_f + B \times L_o \times F_b
\end{array}
\end{equation}
where \(B\) is the batch size, \(L_i\) is the input sentence length, \(L_o\) is the output sentence length, \(F_f\) denotes the FLOPs for the forward pass of one token, and \(F_b\) represents the FLOPs for the backward pass, which generally requires approximately twice the FLOPs of the forward pass.
In the case of SLAP, which incorporates pruning, the computational requirement is reduced by a factor corresponding to the pruning rate \(\alpha\). The computation for SLAP can be expressed as:
\begin{equation}
\begin{array}{l}
\text{SLAP FLOPs} = B \times (L_i + L_o) \times F_f + (1 - \alpha) \times B \times L_o \times F_b
\end{array}
\end{equation}
The backward pass computation is scaled down by \((1 - \alpha)\). For example, with an 70\% pruning rate, only 30\% of the backward pass computations need to be performed.
The comparison of the computational requirements of SLAP, Random, and Full Batch methods at a 70\% pruning rate is illustrated in ~\autoref{fig:d2}. It is evident that our method can reduce computational requirements by at least 30\% while maintaining equivalent loss. This demonstrates the efficiency of SLAP in significantly lowering the computation burden without compromising the performance.


\section{Conclusion}
In conclusion, our proposed method \textbf{SLAP}, significantly enhances the efficiency of instruction tuning for LLMs by focusing on the learnability of whole batch data rather than individual samples. By employing distribution-aware stratified sampling to ensure data distribution coverage and maximizing the relative distance between batch samples for diversity, SLAP effectively curates high-quality, diverse data. Additionally, it utilizes Hessian-approximated gradient optimization to guide batch selection strategy, leading to superior performance compared to previous state-of-the-art methods. Our experiments demonstrate that SLAP achieves robust generalization across various downstream tasks and models, reduces computational cost by 20-40\%, and consistently delivers optimal results in multilingual translation, QA, and multi-dialogue evaluations.
%
%
%
%

\bibliographystyle{splncs04} 
\bibliography{reference}   

\end{document}